\documentclass{nature}

\usepackage{amssymb}
\usepackage{graphicx}
\usepackage{times}
\usepackage{epsfig}
\usepackage{amsmath}
\usepackage{subfigure}

\usepackage{mathtools}
\usepackage{epstopdf}
\usepackage[font=scriptsize]{caption}
\usepackage{float}
\usepackage[]{algorithm2e}
\newcommand{\argmax}{\arg\!\max}

\title{A Random-Fern based Feature Approach for Image Matching}

\author{Yong Khoo, Seo-hyeon Keun}

\begin{document}

\maketitle

\begin{abstract}
Image or object recognition is an important task in computer vision. With the hight-speed processing power on modern platforms and the availability of mobile phones everywhere, millions of photos are uploaded to the internet per minute, it is critical to establish a generic framework for fast and accurate image processing for automatic recognition and information retrieval. In this paper, we proposed an efficient image recognition and matching method that is originally derived from Naive Bayesian classification method to construct a probabilistic model. Our method support real-time performance and have very high ability to distinguish similar images with high details. Experiments are conducted together with intensive comparison with state-of-the-arts on image matching, such as Ferns recognition and SIFT recognition. The results demonstrate satisfactory performance. 
\end{abstract}

\section*{Introduction}
Recognition is one of hot topics in computer vision \cite{iref1}. The main idea of identification of images is finding out the critical and symbolic correspondences among them \cite{iref2}. The more these correspondences the images have, the more evidences to put them in the same class. Nowadays, most of 3D animation tools enable people to create a skeleton of a 3D model manually.
Although this technique provides great help to 3D animation, especially some strange 3D objects,
it is not friendly enough to novices and time-consuming when a large number of 3D characters are
in need. Apart from creating a new easy-learn and convenient system for 3D animation, automatic
generation of skeletons for 3D object is also a choice. In the first paper,Wade and Parent1 introduce
an algorithm for automated construction of skeleton from mesh data. In the second paper, Baran
and Popovic create a system, Pinocchio, that construct a skeleton and embed it into the characters
for the use of animation.

In this paper, I am going to make a simple illustration on Ferns recognition based on 's work. Features matching is one of tools on image recognition. However, a large number of features matching does not represent a good result, and they also exhaust time and memory. Ferns are groups of random features and this idea manage to solve the problems above.\\
The researchers apply a Naive Bayesian classifier to find the probability on a training set of distinct classes of images using Ferns. Then, the classifier use Ferns to calculate the probability of a test image. Comparing the probability of the test image and that of the training images, the researchers can find which class the test image in.

\section*{Related Works}
Generally speaking, there are two major methods on recognition in computer vision.\\
The first way uses local descriptors that is unchangeable in the deformations of reference image. In 1997, Schmid and Mohr\cite{iref3} used local greyvalue invariants to find the objective image from large image databases. In 2001, Lowe\cite{iref4} presented a 3D Object recognition using combination of features from different views. In 2004, Lowe\cite{iref6} achieved image recognition for different poses of an object using the features that are not influenced by image scale and rotation.\\
The second method uses a large number of data on a probability model to find the high possible existing patches in distortion of the original image. The distorted images containing huge existences of these patches can be thought as a class \cite{iref5}. In 2006, Fei-Fei et al.\cite{iref8} implemented a Bayesian on probabilistic models to learn about new object categories from previous learned categories \cite{iref7}. In the same year, Lepetit and Fua\cite{iref10} used randomized tree\cite{iref12} to create a system for object detection.

\section*{Method}
This section explains the idea of Ferns in the following paragraphs. First of all, the researchers find the keypoints in a model image. I will come back to talk about the model image after demonstration of the approach. The revised Naïve Bayes Classifier is employed in Ferns. The formula, calculating the probability of the class given the features, is displayed below,
$$\hat{c_i}=\argmax_{c_i} P(C=c_i|f_1,f_2,...,f_N),$$
where $c_i, i=1,...,H$ represents the set of classes and $f_j, j=1,...,N$ means the set of binary features.\\
A class is defined as the set of all possible appearances of the image patch surrounding a keypoint. For further understanding of this defintion, I take an image of a dog as an example. How can a computer know another image containing the same dog or say they are in the same class? Both images should have many common keypoints \cite{iref9}, and each image patch surrounding the common keypont are matched. The number of pairs of correspondent keypoints is large enough to prove that two images should be placed in the same class.\\
The binary features are located in the image patches and are used for classification. The binary features are calculated under the following function,
$$f_j= \begin{cases}
    1	       & \quad \text{if } I(\mathbf{d}_{j,1})<I(\mathbf{d}_{j,2})\\
    0  & \quad \text{otherwise}\\
  \end{cases}$$
where $I$ means the image patch, and $\mathbf{d}_{j,1}$ and  $\mathbf{d}_{j,2}$ means two pixel locations in the image patch. The sum of features in all the image patches are $N$. The number N is required to be large enough for accurate classification. According to Bayes' Formula,
$$ P(C=c_i|f_1,f_2,...,f_N)=\frac{P(f_1,f_2,...,f_N|C=c_i)P(C=c_i)}{P(f_1,f_2,...,f_N)}, $$
then, the researchers yields,
$$\hat{c_i}=\argmax_{c_i} P(f_1,f_2,...,f_N|C=c_i).$$
There are $N$ features. A value of a feature has to be 0 or 1. If we put all the values in a line in the ascending index order, we will get a binary value. There are $2^N$ situations, and we can convert the binary value into the decimal value. The decimal value is range from 0 to $2^N$. Then, each class would store $2^N$ entries, which is not feasible. However, assuming features are complete independent, the representation of the joint probability is compressed to this,
$$P(f_1,f_2,...,f_N|C=c_i)=\prod_{j=1}^NP(f_j|C=c_i).$$
Because the correlation between features should be totally neglected, the researchers seperate all the features into $M$ groups in the same size $S$, $S=\frac{N}{M}$. The researchers define these groups as Ferns. The conditional probability function for features in each Fern becomes,
$$P(f_1,f_2,...,f_N|C=c_i)=\prod_{k=1}^MP(F_k|C=c_i),$$
where $M$ represents the total number of Ferns and the total groups of features. $$F_k=\{f_{\sigma(k,1)},f_{\sigma(k,2)},...,f_{\sigma(k,S)}\},k=1,...,M$$ $F_k$ represents the $k^{th}$ fern and $\sigma(k,j)$ is a function that makes features in a random order. The researchers put the features in a random order and pick the first $S$ number in $F_1$, pick the second $S$ number in $F_2$... $S=11$, $M$ is range from 30 to 50, and $N$ is about 450, because, after multiple practices, the researchers think these settings can give a good results. More, these settings can change with the consideration of trade-offs between performance and memory.\\
With preparation of formula, the researchers begin training for Ferns. They use model images as classes and detects  a group of keypoints for each class. For each model image, they deform it and find its keypoints again and again. Given that the training images are deform from the model images, the model images and the correspondent deformations are grouped into the same class. Suppose a model image has 1000 deformations, each deformations has it own set of keypoints and the number of keypoints in each set may be different. The researchers picked the keypoints which most commonly appear in the deformation images from the model image into a set. For example, if there is a keypoint detected in the model image and appears in every deformation, this keypoint should be selected \cite{iref11}. The set of these keypoints is assigned a unique class number.\\
Now, we get the class conditional probability for each Fern $F_m$,
$$p_{k,c_i}=P(F_m=k|C=c_i),$$
where $k=1,2,...,2^S$. Thus, the value of a fern is range from 1 to $2^S$. The sum of the probability for each Fern is 1.
$$\sum^K_{k=1}p_{k,c_{i}}=1.$$
Another approach to get $p_{k,c_{i}}$ is,
$$p_{k,c_{i}}=\frac{N_{k,c_{i}}}{N_{c_{i}}},$$
where $N_{k,c_{i}}$ is the number of training sample with the value of the $k^{th}$ Fern in class $c_i$ and $N_{c_{i}}$ is the total number of samples in class $c_i$. However, $p_{k,c_{i}}$ may be zero, because some Fern values is not reachable for some class. For example, there is a dog image. Its Ferns' values only have 1,2,3, and 4. We know the value of Fern is range from 1 to $2^S$, and $S=11$, $N_{k,c_{dog}}=0$, where $k>4$. Since we use multiplication in the conditional probability function, one zero can lead to a bad result. Therefore, the researchers introduce a regularization term $N_{r}$ into $p_{k,c_{i}}$,
$$p_{k,c_{i}}=\frac{N_{k,c_{i}}+N_r}{N_{c_{i}}+K\times N_r}.$$
According to all the researchers experiments, $N_r=1$ provide a good result.\\
Finally, we get this formula,
$$\hat{c_i}=\argmax_{c_i}\prod_{k=1}^M \frac{N_{k,c_{i}}+N_r}{N_{c_{i}}+K\times N_r}.$$

\section*{Experiments and result}
The researchers carry out some experiments on Ferns and compare it to SIFT.
\subsection{Ferns vs SIFT to detect planar objects}
\begin{figure*}[h]
\centerline{\epsfig{figure=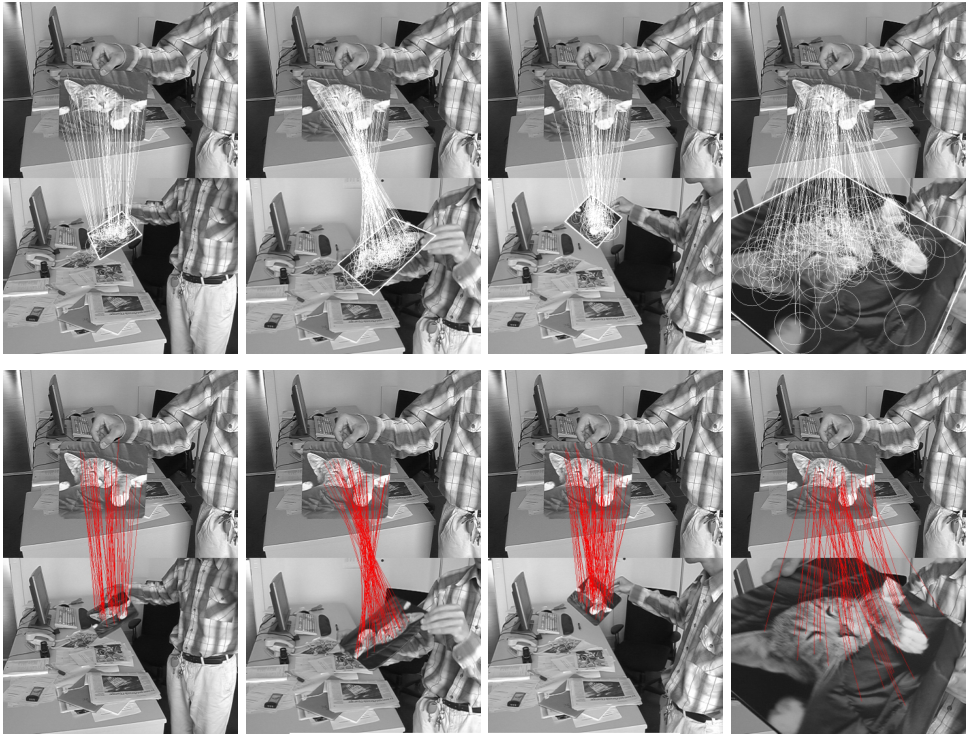,width=14cm}}
\caption{Mouse pat matching. The reference images are at the top of each row images and a few video frame are at the bottom of each row images}
\end{figure*}
\begin{figure*}[h]
\centerline{\epsfig{figure=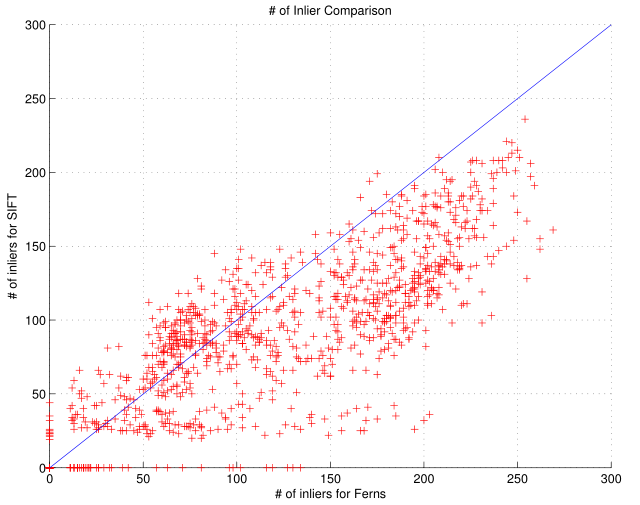,width=14cm}}
\caption{Scatter plot showing the number of inliers for each frame.}
\end{figure*}

The experiments and result are displayed in Figure 1 and Figure 2. The top images on each rows in Figure 1 are reference images. The bottom images on each rows in Figure 1 are sample frames from a video. The researchers test the matching from the mouse pad in the reference images to the mouse pad in the video. Since they use video, it is easy for them to perform a rotations, scalings and deformations for the test images. Figure 2 is a result. It shows that Ferns has more inliers than SIFT, so Ferns has a better result.

\subsection{Ferns vs SIFT to detect 3D objects}
In this experiments, different views of four 3D objects are used to generate training images and tests images. Figure 3 show the images of the objects.
\begin{figure*}[h]
\centerline{\epsfig{figure=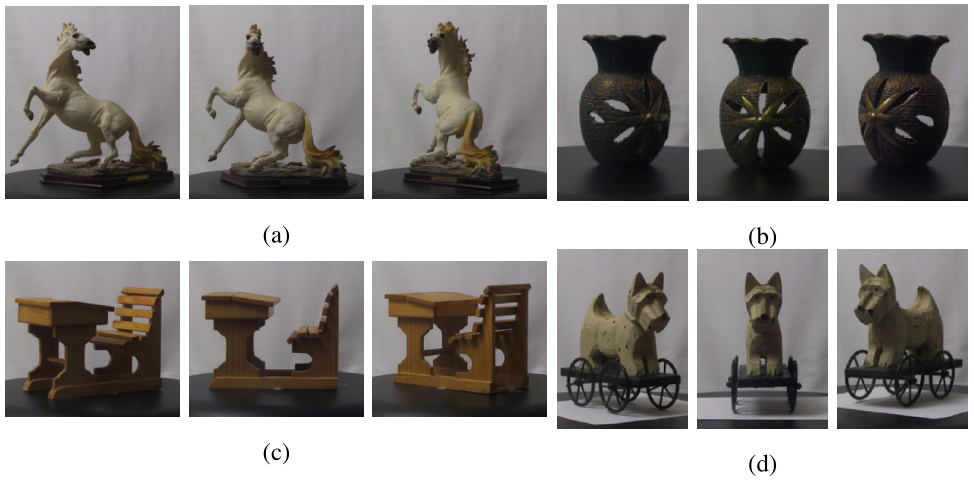,width=14cm}}
\caption{Four 3D objects for experiments. a) horse, b) vase, c) desk, d)dog}
\end{figure*}
Figure 4 and Figure 5 show how the researchers obtain the data. 
\begin{figure*}[h]
\centerline{\epsfig{figure=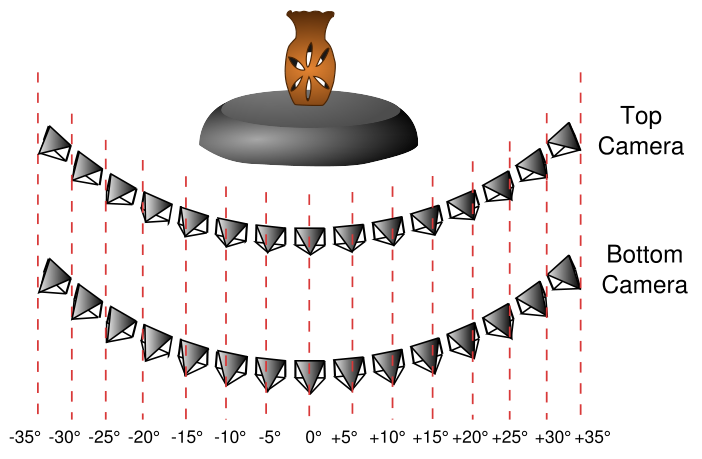,width=14cm}}
\caption{Take vase as an example to show the way to get ground truth data for 3D object,}
\end{figure*}
\begin{figure*}[h]
\centerline{\epsfig{figure=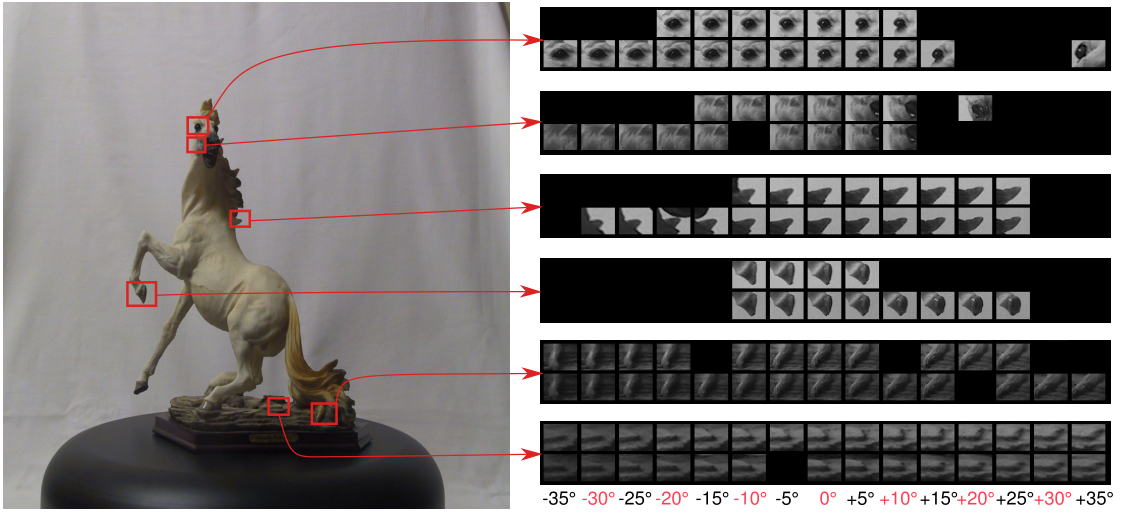,width=14cm}}
\caption{Detect 6 keypoints on horse and get the ground truth data for training and testing}
\end{figure*}
In Figure 4, the researchers catch ground truth data by 3D objection detection from top and bottom camera in multiple views. In Figure 5, the researchers find six keypoints in the horse and obtain the group truth data of them. Then, they use the degrees in red as training data while those in black as testing data. 
\begin{figure*}[h]
\centerline{\epsfig{figure=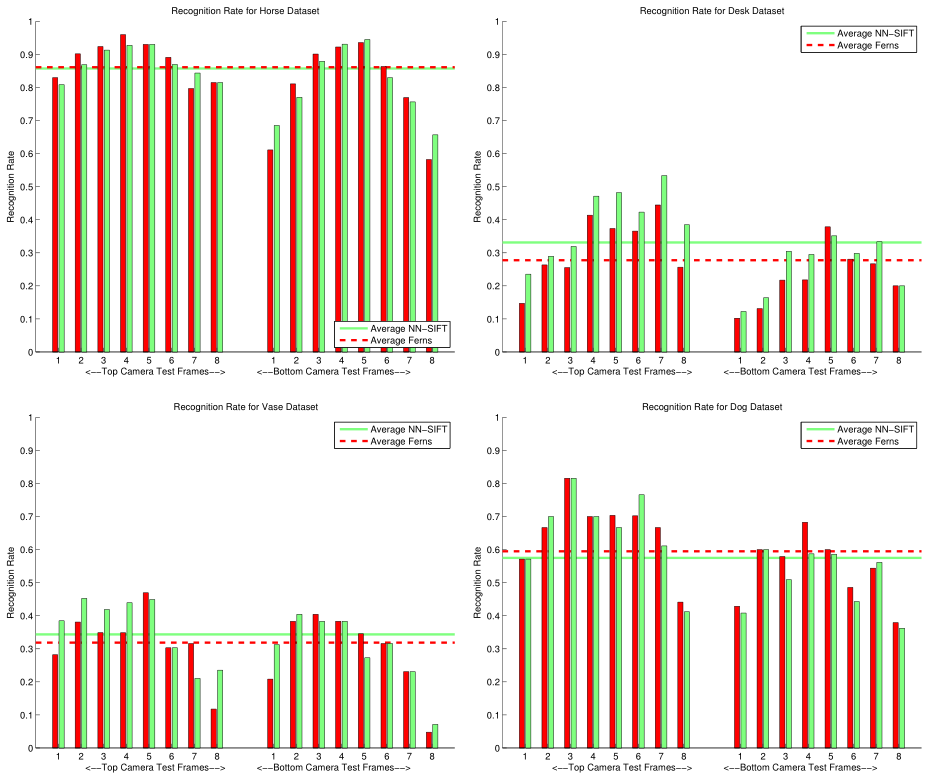,width=14cm}}
\caption{recognition rates for 3D objects. The red bar on the left means the rate for Ferns and the green on means that for SIFT}
\end{figure*}
Figure 6 show result of four 3D objects in both method. It is hard to conclude which one is better from the result. However, if the training frames increases, SIFT mathcing will slow down while Ferns will not. This advantage is result from the flexibility of the size of Ferns.


\section*{Reference}

\begin{thebibliography}{1}

\bibitem{iref1} Ozuysal, Mustafa: Fast keypoint recognition using random ferns. IEEE transactions on pattern analysis and machine intelligence 32.3 (2010): 448-461.

\bibitem{iref2} Shen, J., Cheung, S. :Layer depth denoising and completion for structured-light rgb-d cameras. IEEE Conference on Computer Vision and Pattern Recognition (CVPR), 1187-1194, 2013.


\bibitem{iref3} Schmid, Cordelia, and Roger Mohr.: Local grayvalue invariants for image retrieval. IEEE transactions on pattern analysis and machine intelligence 19.5 (1997): 530-535.

\bibitem{iref4} Lowe, David G.: Local feature view clustering for 3D object recognition. Computer Vision and Pattern Recognition, 2001. CVPR 2001. Proceedings of the 2001 IEEE Computer Society Conference on. Vol. 1. IEEE, 2001.

\bibitem{iref5} Shen, Ju, Su, P.-C, Cheung, S.-C, Zhao, J.: Virtual mirror rendering with stationary rgb-d cameras and stored 3-d background. IEEE Transactions on Image Processing, vol 22, 3433-3448, 2013.

\bibitem{iref6} Lowe, David G.: Distinctive image features from scale-invariant keypoints. International journal of computer vision 60.2 (2004): 91-110.

\bibitem{iref7} Shen, J., Yang, J., Taha-abusneineh, S., Payne, B., Hitz, M.: Structure Preserving Large Imagery Reconstruction, Journal of Cyber Security and Mobility, Vol. 3  Issue. 3,  2014.

\bibitem{iref8} Fei-Fei, Li, Rob Fergus, and Pietro Perona.: One-shot learning of object categories. IEEE transactions on pattern analysis and machine intelligence 28.4 (2006): 594-611.


\bibitem{iref9} Shen, J., Tan, W.: Image-based indoor place-finder using image to plane matching,  IEEE International Conference on Multimedia and Expo (ICME), 2013.

\bibitem{iref10} Lepetit, Vincent, and Pascal Fua.: Keypoint recognition using randomized trees. IEEE transactions on pattern analysis and machine intelligence 28.9 (2006): 1465-1479.


\bibitem{iref11} Jianjun Yang,Yin Wang, Honggang Wang, Kun Hua, Wei Wang and Ju Shen, Automatic Objects Removal for Scene Completion, the 33rd Annual IEEE International Conference on Computer Communications (INFOCOM'14), Workshop on Security and Privacy in Big Data, 2014.

\bibitem{iref12} Amit, Yali, and Donald Geman.: Shape quantization and recognition with randomized trees. Neural computation 9.7 (1997): 1545-1588.



\end{thebibliography}
\end{document}